\documentclass{article}

\usepackage{arxiv}

\usepackage[utf8]{inputenc} % allow utf-8 input
\usepackage[T1]{fontenc}    % use 8-bit T1 fonts
\usepackage{hyperref}       % hyperlinks
\usepackage{url}            % simple URL typesetting
\usepackage{booktabs}       % professional-quality tables
\usepackage{amsfonts}       % blackboard math symbols
\usepackage{nicefrac}       % compact symbols for 1/2, etc.
\usepackage{microtype}      % microtypography
\usepackage{lipsum}		% Can be removed after putting your text content
\usepackage{graphicx}
\usepackage{natbib}
\usepackage{doi}
\usepackage{multirow}
\usepackage{amsmath}

\title{Enhancing Sindhi Word Segmentation using Subword Representation Learning and
Position-aware Self-attention}

\author{ \href{https://orcid.org/0000-0002-9392-459X}{\includegraphics[scale=0.06]{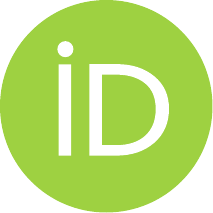}\hspace{1mm}Wazir Ali}\thanks{Corresponding author (Wazir Ali, aliwazirjam@gmail.com)} \\
	Department of Computer Science\\
	Institute of Business Management\\
	Karachi, PK 75190 \\
	\texttt{aliwazirjam@gmail.com} \\
	\And
 Jay Kumar\\
 Faculty of Computer Science\\
	Dalhousie University, Halifax -  NS, Canada\\ 
	\texttt{jay.kumar@dal.ca} \\
 \And
	Saifullah Tumrani\\
	Heidelberg University, Germany\\ 
 Bahria University, Pakistan\\
	\texttt{saifullahtumrani.bukc@bahria.edu.pk} \\
 \And
 Redhwan Nour\\
 College of Computer Science and Engineering\\
 Taibah University, Saudi Arabia\\
	\texttt{rnour@taibahu.edu.sa} \\
 \And
 Adeeb Noor\\
	Faculty of Computing and Information Technology\\
	King Abdulaziz University, Saudi Arabia\\
	\texttt{arnoor@kau.edu.sa} \\
 \And
 Zenglin Xu\\
	School of Computer Science and Technology\\
	Harbin Institute of Technology, Shenzhen 510085, China\\
	\texttt{zenglin@gmail.com} \\
}

\date{}

\renewcommand{\shorttitle}{\textit{arXiv} Template}

\hypersetup{
pdftitle={A template for the arxiv style},
pdfsubject={q-bio.NC, q-bio.QM},
pdfauthor={David S.~Hippocampus, Elias D.~Striatum},
pdfkeywords={First keyword, Second keyword, More},
}

\begin{document}
\maketitle

\begin{abstract}
 Sindhi word segmentation is a challenging task due to space omission and insertion issues. The Sindhi language itself adds to this complexity. It's cursive and consists of characters with inherent joining and non-joining properties, independent of word boundaries. Existing Sindhi word segmentation methods rely on designing and combining hand-crafted features. However, these methods have limitations, such as difficulty handling out-of-vocabulary words, limited robustness for other languages, and inefficiency with large amounts of noisy or raw text. Neural network-based models, in contrast, can automatically capture word boundary information without requiring prior knowledge.
In this paper, we propose a Subword-Guided Neural Word Segmenter (SGNWS) that addresses word segmentation as a sequence labeling task. The SGNWS model incorporates subword representation learning through a bidirectional long short-term memory encoder, position-aware self-attention, and a conditional random field. Our empirical results demonstrate that the SGNWS model achieves state-of-the-art performance in Sindhi word segmentation on six datasets. 
\end{abstract}

% keywords can be removed

\section{Introduction}
The word segmentation is a fundamental and challenging task in Sindhi text classification and other Natural Language Processing (NLP) tasks~\cite{bhatti2014word}. A word segmenter determines the word boundaries in the shape of beginning and ending~\cite{ding2016word}. It has been largely investigated in many space-delimited languages including English~\cite{polka2012word,wang2016morphological}, Arabic~\cite{almuhareb2019arabic}, Urdu~\cite{zia2018urdu} and non-space delimited languages including Chinese~\cite{li2023markbert}, Japanese~\cite{kitagawa2018long}, and Burmese~\cite{Zhang:2018:WSB:3292011.3232537}. However, the word segmentation in low-resource Sindhi language has not been studied well~\cite{jamro2017sindhi}, mainly due to the lack of language resources and its morphological complexities. 

Sindhi word segmentation is a challenging task because it exhibits space omission and space insertion problems~\cite{mahar2012model}. This is partly because the Arabic script, which is cursive in nature, consists of characters that have inherent joining and non-joining attributes regardless of a word boundary~\cite{bhatti2014word}. Apart from the discussed problems, there is no gold-standard benchmark corpus for Sindhi to evaluate the segmentation task~\cite{jamro2017sindhi}. In summary, the SWS task is difficult, important, and not studied as a sequence modeling problem. The previous approaches mainly rely on rule-based methods~\cite{mahar2011algorithms,mahar2012model,bhatti2014word,dootio2017automatic}, which have certain limitations such as the inability to deal with Out-Of-Vocabulary (OOV) words, less robustness for other languages, and the algorithms' inefficiency to deal with a large amount of noisy text. Thus, the existing approaches lack the robustness and applicability towards open-source implementation.

Neural networks have gained great attention in the NLP community~\cite{zaman2024leveraging,al2024deberta,murat2024low}. On the one hand, state-of-the-art sequence tagging systems rely heavily on large amounts of task-specific knowledge~\cite{ma2016end}. This knowledge comes in the form of hand-engineered features and data pre-processing. While neural network models have been extensively investigated for separate word segmentation in many space-delimited and non-space-delimited languages, Sindhi word segmentation has not yet been addressed as a sequence tagging problem using neural network models.

% Although neural network models for separate word segmentation and POS tagging have been intensively investigated, only few work has been aware of neural joint model of the twotasks. There are several exceptions [24]–[27]. Only Kurita et al. (2017) [26] and Shao et al. (2017) [27] have achieved better
% performances than traditional statistical models on a widelyadopted standard CTB5 dataset . The former requires heavy feature engineering, designing more than 40 atomic features to construct their neural network structure, and the later exploits
% a CRF-based sequence labeling framework, resulting much
% lower performances than the former one

%On the other hand, we can enhance the performance of NN based models by incorporating unsupervised neural representations~\cite{bojanowski2017enriching}. More recent state-of-the-art approaches concatenate up to four distinct types of word representations including classical~\cite{mikolov2013distributed}, character-level~\cite{bojanowski2017enriching}, deep contextualized~\cite{peters2017semi},~\cite{akbik2018contextual} and task-oriented~\cite{liu2018task}.

Recently, RNN variants, including LSTMs and BiLSTMs, have been successfully exploited for sequence tagging problems~\cite{rai2023deep,kilicc2023keyword,zaman2024leveraging}. However, LSTMs are limited in that they can only encode sequences in a unidirectional way. This limitation is addressed by BiLSTMs, which stack two LSTMs to create a bidirectional encoder~\cite{schuster1997bidirectional}. The BiLSTMs employ a simultaneous training strategy in both positive and negative directions. This is an ideal solution because the bidirectional network can access context from both the right and left directions in sequence labeling, leading to a better understanding of text sequences.

% On the one hand, state-of-the-art sequence tagging systems rely on the large amounts of task specific knowledge~\cite{ma2016end} in the form of hand-engineered features and data pre-processing. On the other hand, we can enhance the performance of NN based models by incorporating unsupervised neural representations~\cite{bojanowski2017enriching}. More recent state-of-the-art approaches concatenate up to four distinct types of word embeddings including classical~\cite{mikolov2013distributed}, character-level~\cite{bojanowski2017enriching}, deep contextualized~\cite{peters2017semi},~\cite{akbik2018contextual} and task-oriented~\cite{liu2018task}. The classical approach efficiently captures the semantic, syntactic word and phrase similarities. While, character-level features, trained on task-specific subword features and the deep context-aware embeddings capture word semantics in context to address the polysemous and context-dependent nature of words. We utilize the task-oriented word embeddings~\cite{liu2018task} with SRL strategy~\cite{labeau2017character}.

%
We propose a language-independent SGNWS model that incorporates neural model with a powerful BiLSTM encoder, subword representation learning (SRL), and self-attention mechanism. We convert word segmentation into a sequence tagging problem. To the best of our knowledge, this is the first attempt to tackle SWS as a sequence labeling task. Our novel contributions are listed as follows:
\begin{table}
    \centering
    \caption{Sequence tagging scheme by assigning the character-level [B, I, E, S, X] tags to unlabelled corpus for the word boundary detection}
    
    \label{tab:tagging_scheme}
    \begin{tabular}{c|l}
    \hline
     B & \textbf{Beginning} of current character of a word consisting\\& of more than one character\\
     I & \textbf{Inside} or middle of a current character of a word \\& consisting  of more than two characters. \\
     E & \textbf{Ending} denotes that the current character is the end \\& of a word consisting of more than one character. \\
     S & \textbf{Single} is the current character a word consisting  of \\& only one character. \\
     X & Hard space between words, character, symbols etc.,\\
     \hline
    \end{tabular}
\end{table}

\begin{itemize}
\item We treat Sindhi Word Segmentation (SWS) as a sequence tagging problem, assigning character-level [B, I, E, S, X] tags to an unlabeled corpus for word boundary detection. 
\item We incorporate position-aware self-attention to capture the positional information of the given sequence. This allows the attention mechanism to focus on different parts of the input sequence based on their relevance to the task.
\item  The proposed SGNWS model eliminates the need for highly complex, hand-engineered or rule-based word segmentation algorithms by adopting BiLSTM encoder, SRL, and self-attention mechansim.
\end{itemize}

\section{Related work}
\label{sec:relatedwork}
The word segmentation has been well studied in English~\cite{polka2012word,wang2016morphological}, Arabic~\cite{almuhareb2019arabic}, Chinese~\cite{duan2020attention},  and many other languages. Little work has been carried out to address the word segmentation problem for Sindhi. Presently, Sindhi language is being written in two famous writing system of Persion-Arabic and Devanagari~\cite{motlani2016developing}. However, Persian-Arabic is the standard script of Sindhi~\cite{mahar2012model}, which contains rich morphology due to the frequent usage of prefixes and suffixes to express inflections and derivations, which makes it complex morphological language.

The existing SWS models rely on the rule-based approaches~\cite{jamro2017sindhi}. Initially,~\cite{mahar2011algorithms} coined the SWS problem by introducing the first word segmentation model using several rule-based algorithms. The proposed model was evaluated on a small dataset 16,601 lexicon with segmentation error rate (SER) of 9.54$\%$. Later,~\cite{mahar2012model} proposed another rule-based model with $91.76\%$ segmentation accuracy. They performed the word segmentation in three steps: the first step consists of input and segmentation with white space,  second step is used for the segmentation of simple and compound words, while the third step deals with the segmentation of complex words. Moreover,~\cite{bhatti2014word} proposed a SWS model by evaluating the dataset of 157K tokens obtained from news corpus and dictionary lexicon. Their proposed model achieves good performance on the dictionary lexicon, but poorly performed in dealing with news and books corpus.~\cite{dootio2017automatic} proposed two algorithms for Sindhi lexicon lemmatization and stemming. The rule-based methods have certain limitations such as inability to deal with out of vocabulary words, less robustness for other datasets or languages, and the algorithms inefficiency to deal with a large amount of noisy or raw text.

Recently, neural sequence to sequence models including RNNs have largely gained popularity~\cite{young2018recent} by greatly simplifying the learning and decoding process in a number of NLP applications~\cite{collobert2011natural,ma2016end,yang2017neural,yao2016bi} including word segmentation~\cite{almuhareb2019arabic,shao2017cross}. The BiLSTM~\cite{schuster1997bidirectional} is an ideal solution to learn the sequences in a language because it has access to both the contexts of right and left directions. The BiLSTM network has been largely employed for word segmentation in Chinese~\cite{chen2015long,shao2017character}, Japanese~\cite{kitagawa2018long} and Arabic~\cite{almuhareb2019arabic}, and other languages~\cite{yang2017neural,yao2016bi,ma2018state} by achieving excellent performance without relying on any external feature engineering strategies. On the one hand, state-of-the-art sequence tagging systems rely on large amounts of task-specific knowledge~\cite{ma2016end} in the form of hand-engineered features and data pre-processing. On the other hand, the performance of neural models rely on the incorporating unsupervised neural representation learning including classical~\cite{mikolov2013distributed}, character-level~\cite{bojanowski2017enriching},  contextualized~\cite{akbik2018contextual} and task-oriented~\cite{liu2018task}. More recently, the integration of self-attention mechanism~\cite{vaswani2017attention} in neural models has also yielded new state-of-the-art results word segmentation~\cite{duan2020attention}. Moreover, the last layer of neural models has a significant impact on the performance. The CRF~\cite{lafferty2001conditional} is broadly used in the sequence tagging tasks~\cite{kitagawa2018long,ma2016end,huang2015bidirectional,bhumireddypalli2023enhanced} for decoding in neural models. Taking advantage of language-independent neural models for SWS, we propose a model that efficiently captures the character-level information with SRL by converting the segmentation into a sequence tagging problem.

\section{Sindhi Morphology}
\label{sec:sindhi_writingsystem}
The Persian-Arabic is a standard writing script for Sindhi, which is cursive and written from right to left direction~\cite{narejo2016morphology}. It contains rich morphology~\cite{mahar2012model} due to the frequent usage of prefixes and suffixes to express inflections and derivations, which makes it a complex morphological language. The alphabet of Sindhi Persian-Arabic consists of 52 basic letters, 29 derived from Arabic language, 03 from the Persian language, and 20 modified letters ~\cite{jamro2017sindhi}. It also uses 03 secondary letters, 07  honorific symbols and diacritic marks~\cite{narejo2016morphology,rahman2010towards}. Interestingly, the shape of some letters in Sindhi change the form according to their position in a word~\cite{bhatti2014word}, such letters are referred as \textit{joiners}. Thus, a \textit{joiner} have at most four shapes; i) initial ii) middle iii) final and iv) isolated, as Table \ref{tab:sindhi_character_shapes} depicts an example of some letters.  Whereas the position-independent letters having final or isolated form are referred as \textit{non-joiners}. Specifically, white spaces are used to detect word boundaries in Sindhi. However, writers omit a hard space between two words. Therefore, a phrase or a sentence that ends with non-joiner letters becomes one token. In the first case, the words are joined with their preceding and succeeding words in the absence of white space, which leads to misspellings. In the second case, the shape of characters remains identical even in the absence of white space. Due to position-independent and space-independent letters, the SWS exhibits both challenges of space insertion and space omission~\cite{mahar2011algorithms,rahman2010towards}. 
\begingroup
\setlength{\tabcolsep}{8pt} % Default value: 6pt
\renewcommand{\arraystretch}{1.4} % Default value: 1
\begin{table}
	\centering
	\caption{Various shapes of Sindhi alphabet according to their position in words. Roman transliteration of every isolated letter is given for the ease of reading.}
   \begin{tabular}{ccccl}
    \hline
    Ending & Middle & Initial & Isolated & Roman\\
    \hline
    {\includegraphics[width=0.025\textwidth]{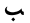} } &  {\includegraphics[width=0.015\textwidth]{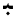} } & 
    {\includegraphics[width=0.015\textwidth]{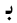} } & 
    {\includegraphics[width=0.025\textwidth]{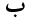} } & 
    B$\bar{e}$ \\
    
    {\includegraphics[width=0.018\textwidth]{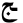} } &  {\includegraphics[width=0.020\textwidth]{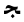} } & 
    {\includegraphics[width=0.018\textwidth]{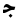} } & 
    {\includegraphics[width=0.015\textwidth]{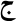} } & 
    J$\bar{i}$m \\
    
    {\includegraphics[width=0.025\textwidth]{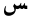} } &  {\includegraphics[width=0.020\textwidth]{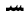} } & 
    {\includegraphics[width=0.018\textwidth]{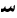} } & 
    {\includegraphics[width=0.025\textwidth]{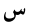} } & 
    S$\bar{i}$n \\
    
    {\includegraphics[width=0.014\textwidth]{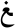} } &  {\includegraphics[width=0.018\textwidth]{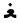} } & 
    {\includegraphics[width=0.018\textwidth]{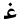} } & 
    {\includegraphics[width=0.015\textwidth]{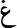} } & 
    $\breve{g}$ain \\
    
    {\includegraphics[width=0.022\textwidth]{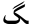} } &  {\includegraphics[width=0.018\textwidth]{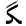} } & 
    {\includegraphics[width=0.018\textwidth]{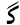} } & 
    {\includegraphics[width=0.020\textwidth]{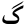} } & 
    G$\bar{a}$f \\
    
    {\includegraphics[width=0.020\textwidth]{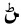} } &  {\includegraphics[width=0.018\textwidth]{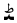} } & 
    {\includegraphics[width=0.018\textwidth]{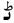} } & 
    {\includegraphics[width=0.018\textwidth]{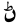} } & 
    N$\breve{u}$n \\
    \hline
    \end{tabular}
    \label{tab:sindhi_character_shapes}
\end{table}
\endgroup
\begin{table*}[t!]
	\centering
	\caption{Complete list of Sindhi joiner and non-joiner letters, (i) denote joiner letters (ii) non-joiners, and (iii) non-joiner secondary letters.}
	\includegraphics[width=0.98\textwidth]{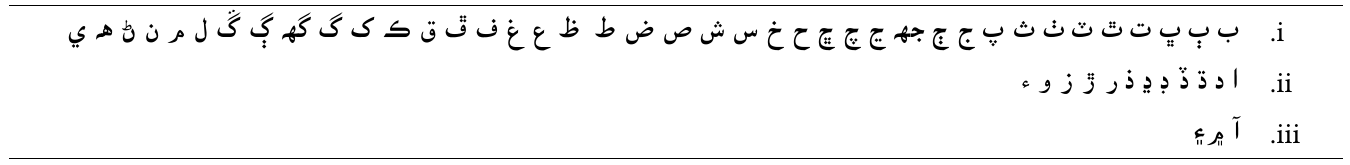}
	\label{tab:sindhi_joiner_nonjoiner_words}
\end{table*}
\subsection{Space Omission}
\label{space_omission}
The space omission is a common phenomenon in Sindhi words that end with the non-joiner letters. However, the absence of white space exhibit the correct shape of words such as Table \ref{tab:sindhi_joiner_nonjoiner_words} shows an example of a Sindhi sentence with and without the use of white space. But computationally, that sentence consists of one token without the use of white spaces between words. Whereas the sentence consists of eight tokens with the use of white space between words. Therefore, the omission of white space between words ending with non-joiner letters raises a computational issue.

\subsection{Space Insertion}
\label{subsec:space_insertion}
Another challenge in SWS arises when combining two or more root words (\textit{morphemes}) form a new standalone single word (see Table \ref{tab:sindhi_words}). In such cases, writers omit white space if the first morpheme ends with a joiner letter. However, white space prevents it's joining with the next morpheme so that the word retains a valid visual form. The missing space insertion leads to the formation of compound words and often misspelling. Hence, white space is essential in this case for the ease of readability and correct spelling of Sindhi words.
%  Moreover, writers may correctly omit the hard space because a word's shape remains identical if the first morpheme ends in a non-joiner.

%
\begin{table}
	\centering
	\caption{An example of a Sindhi sentence, all words end with the non-joiner letters. (i)  denote the words with white space (the tokens are separated with `-' symbol), (ii) without white space (iii) Roman transliteration of Sindhi sentence (iv) is the English translation of a Sindhi sentence.}
    \label{tab:sindhi_joiner_nonjoiner_words}
    \includegraphics[width=0.4\textwidth]{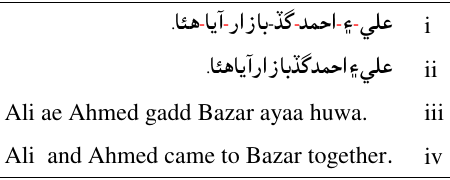}
\end{table}

\begingroup
\setlength{\tabcolsep}{8pt} % Default value: 6pt
\renewcommand{\arraystretch}{1.4} % Default value: 1
\begin{table*}
	\centering
	\caption{Sindhi word types with an example of space insertion, along with English translation. (i) represent the words with white space (`-’ symbol represents space), and (ii) without space. The Roman transliteration is given for ease of reading.}
    \begin{tabular}{lccll}
    \hline
    Word Type & i. & ii. & Roman & English Translation\\
    \hline
    Affix & 
    {\includegraphics[width=0.067\textwidth]{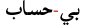} } &  {\includegraphics[width=0.056\textwidth]{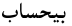} } & Be-hisaab & Uncountable\\
    
    Reduplicate & 
    {\includegraphics[width=0.069\textwidth]{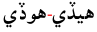} } &  {\includegraphics[width=0.061\textwidth]{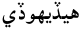} } & haidai hodai & Here and there\\
    
    Compound & 
    {\includegraphics[width=0.077\textwidth]{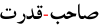} } &  {\includegraphics[width=0.071\textwidth]{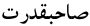} } & saahib-e-Qudrat & Powerful\\
    
    Borrowed & 
    {\includegraphics[width=0.074\textwidth]{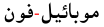} } &  {\includegraphics[width=0.064\textwidth]{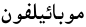} } & Mobile phone & Mobile Phone\\
    
    Abbreviation & 
    {\includegraphics[width=0.079\textwidth]{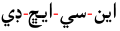} } &  {\includegraphics[width=0.081\textwidth]{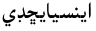} } & Ain Cee Aich Dee & NCHD\\
    \hline
    \end{tabular}
    \label{tab:sindhi_words}
    
\end{table*}
\endgroup
%
% \begin{table}
% 	\centering
% 	\caption{An example of a Sindhi sentence in which all words end with the non-joiner letters. (i)  are the words with white space (the tokens are separated with `-' symbol) after each word. (ii) without white space (iii) Roman transliteration of Sindhi sentence (iv) is the English translation of a Sindhi sentence.}
%     \label{tab:sindhi_joiner_nonjoiner_words}
%     \includegraphics[width=0.7\textwidth]{tab_space_omission.pdf}
% \end{table}
% 

\section{The Proposed Model}
\label{sec:model_architecture}
In this section, we describe the proposed SGNWS model in detail. Firstly, we introduce the tagging scheme and employ language-specific setting based on the morphological complexities, space-omission and space-insertion issues.  we illustrate the neural structures of our SGNWS model, which includes three core parts: the encoder, self-attention, and decoder. Then, the training details of the proposed model are presented.
\begin{figure}[h]
    \centering
    \includegraphics[width=0.55\textwidth]{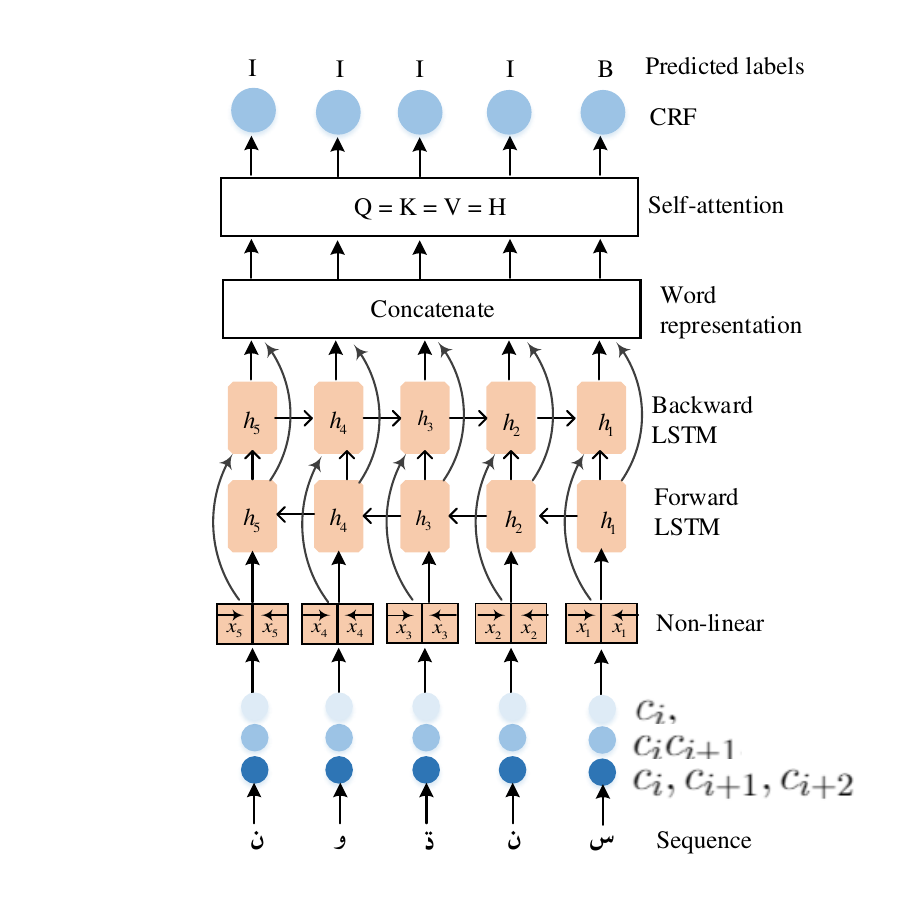}
    \caption{The architecture of the proposed SGNWS model. The input sentences $x_{1}, x_{2}, \dots x_{n}$, are converted into a sequence of character-level subword representations as input. BiLSTM network learn to how to obtain subword features for identifying word boundaries. The output of BiLSTM encoder layer is fed into self-attention layer before decoding. The notation Q=K=V=H signifies that the Query, Key, and Value vectors used in the attention mechanism are all derived from the same hidden layer (H), obtained from the output of a BiLSTM encoder layer. Finally, we employ CRF to obtain the predicted label sequence.}
    \label{fig:proposed_model}
\end{figure}
\subsection{Label set} 
\label{subsec:tagging_scheme} 
We modeled the word segmentation as character-level sequence tagging~\cite{chen2015long}. Theoretically, word boundary can be predicted with binary classification in word segmentation, but in practice, fine-grained tag sets~\cite{zhao2006effective} produce high segmentation accuracy. Following the work~\cite{shao2017character}, we employ four tags [B, I, E, S] to indicate the position of letters at the beginning [B],inside [I], ending [E] of a word, or a single-character/symbol [S], respectively. Additionally, [X] is used to represent the white space to delimit word boundaries. A sentence, as an example of the proposed tagging scheme is depicted in Table \ref{fig:tagging_scheme} by assigning the proposed tags to a sentence.
\begin{table*}
    \centering
    \includegraphics[width=0.8\textwidth]{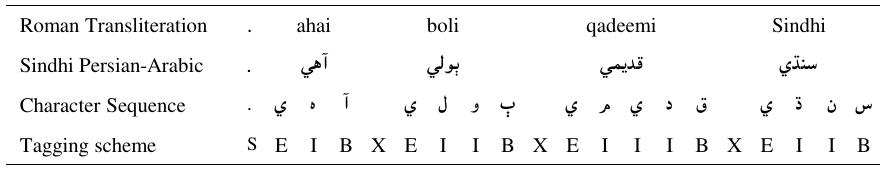}
    \caption{An example of employed character-level sequence tagging scheme for SWS task. The [X] label represents the white spaces. The given Sindhi sentence can be read from right to left, and the Roman transliteration of each Sindhi token can be read from left to right.}
    \label{fig:tagging_scheme}
\end{table*}

\subsection{Language-Specific Setting}
\label{subsec:lang_spec_setting}
Based on the topological factors and baseline models, we employ the following language specific setting.  
\begin{enumerate}
    \item Firstly we separate punctuation marks, then use space-delimited syllables for word boundary detection.
    \item In order to tackle space omission problem, we use SRL based on concatenated trigram representations. 
    \item  We split the large sentences with white-space if the length exceeds more than 300 tokens and do not consider sentences having tokens less than 5. 
    \item The regular hard-space is tagged as [X] in the dataset and punctuation marks are tagged as [S]. However, multi-word tokens without hard-space  such as numerical expressions 689.0967,  date 25-06-2020, money 4736\$, etc., are assigned continuous tags.
    %For example, number 689.0960 is assigned continuous tags of BIIIIIIE.
\end{enumerate}

%Following the majority of sequence tagging models~\cite{shao2017character,zheng2017enhancing,ma2018state}.

\subsection{Encoder}
 The model takes character-level input $x{_t}={c_{1}, c_{2}, c_{3}, \dots c_{i}}$ of character unigrams $c_{i}$, character bigrams $c_{i}, c_{i+1}$, character trigrams $c_{i}, c_{i+1}, c_{i+2}$, and 4-grams of each word $w_{n}$ for SRL, as depicted in Table \ref{tab:sub_word_decomposition}.
 
 %The BiLSTM encoder is built as follows. First,we derive two sequences of input features $\overrightarrow{\mathrm{x}}_{1} \overrightarrow{\mathrm{x}}_{2} \cdots \overrightarrow{\mathrm{x}}_{n}$ and $\overleftarrow{\mathbf{x}}_{1} \overleftarrow{\mathbf{x}}_{2} \ldots \overleftarrow{\mathbf{x}}_{n}$ from the character n-gram level input by neural embedding, respectively. Then, apply a left-to-right LSTM over $\overrightarrow{\mathrm{x}}_{1} \overrightarrow{\mathrm{x}}_{2} \cdots \overrightarrow{\mathrm{x}}_{n}$ and a right-to-left LSTM over$\overleftarrow{\mathbf{x}}_{1} \overleftarrow{\mathbf{x}}_{2} \ldots \overleftarrow{\mathbf{x}}_{n}$ to obtain the bidirectional hidden representations of LSTMs, respectively.

 %Character trigrams have been shown highly effective for Sindhi word segmentation in a number of studies, under the neural setting [19], [26].
   
\subsubsection{Subword Representation Layer}
We make the Subword representations from three types
of discrete sources: character unigram $c_{i}$, bigram ${c_{i}}, {c_{i+1}}$, and trigram $c_{i}, c_{i+1}, c_{i+2}$,  $(i \in[1, n])$ in this work. Formally, we exploit three looking-up tables ${E^{c}}$, ${E^{bc}}$, ${E^{tbc}}$ for
character unigrams, bigrams, and trigrams. Give one character unigram $c_{i}$, its representation $\mathbf{E}_{c_{i}}^{c}$ is obtained by indexing from
$E^{c}$, one character bigram ${c_{i}} {c_{i+1}}$, its representation ${E^{bc}}$, and one trigram $c_{i}, c_{i+1}, c_{i+2}$ its representation ${E^{tbc}}$ is obtained by indexing from ${E^{tbc}}$ accordingly. The three looking-up tables are model parameters which would be
learned during training.
% In the next step, embeddings are fed to the encoder.  We use BiLSTM network~\cite{hochreiter1997long} for SRL by representing each word $w$ from a fixed vocabulary $V$ of unlabeled Sindhi text in a sequence of forward and backward character representations. Such as, character representations $E^{c}=\left[c_{1}, c_{2}, c_{3}, \dots , c_{i}\right]$, bigrams ${E^{B}}=\left[c_{i}, c_{i+1}\right]$, and trigrams ${E^{T}}=\left[c_{i}, c_{i+1}, c_{i+2}\right]$ of a given word are learned to capture the structure of words at morphemic level. Afterwards, we utilize both forward and backward representations by concatenating them:
% \begin{equation}
%     \begin{aligned}
%     \overrightarrow{h}{_{t}}=\text{LSTM}\left(E^{C_{i}}:E^{B_{i}}:E^{T_{i}},\overrightarrow{h}{_{t-1}}\right), \\
%     \overleftarrow{h}{_{t}}=\text{LSTM}\left(E^{C_{i}}:E^{B_{i}}:E^{T_{i}}, \overleftarrow{h}{_{t+1}}\right), \\
%     \text{BiLSTM}\left({Emb_{S}}\right)=\overrightarrow{h}{_{|w|}}: \overleftarrow{h}{_{1}},
%     \label{Eq:concatination_of_bigrams_trigrams}
%     \end{aligned}
% \end{equation}
% where ${Emb_{s}}$ is the concatenated output of Bidirectional $\overrightarrow{h}{_{|w|}}$,  $\overleftarrow{h}{_{1}}$ representations of LSTM layers over the sequence of character n-grams.
%
\begin{table}[!h]
    \centering
    \caption{An example of Sindhi subword decomposition for subword representation learning}
    \includegraphics[width=0.4\textwidth]{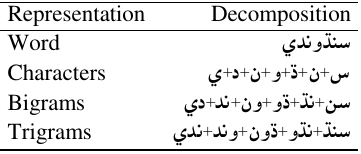}
    \label{tab:sub_word_decomposition}
\end{table}
    
\subsubsection{LSTM Inputs}
In this layer, the BiLSTM encoder takes  input of different neural features to the exploration of character-level unigram, bigram, and trigram based features as inputs. The input representations of both right-to-left and left-to-right at $i^{the}$ position always use the features ending at $c_{i}$, where the last input trigrams right-to-left and the left-to-right LSTMs
are $c_{t} c_{t+1}c_{t+2}$ and $c_{t-2}c_{t-1} c_{t}$, respectively. We employ a non-linear feed-forward neural layer to combine character-level unigrams, bigrams, and bigrams, formalized as follows:

\begin{equation}
    \begin{aligned}
\overrightarrow{{x}}_{t} &=\tanh \left({W}_{c}\left[{E}_{c_{i}}^{c}, {E}_{c_{i-1} c_{i}}^{b c}, {E}_{c_{i-1}{c_{i-2} c_{i}}}^{t b c}\right]+{b}_{c}\right)\\
\overleftarrow{{x}}_{t} &=\tanh \left({W}_{c}\left[{E}_{c_{i}}^{c}, {E}_{c_{i} c_{i+1}}^{t b c}, {E}_{c_{i} c_{i+1} c_{i+2}}^{b c}\right]+{b}_{c}\right),
\end{aligned}
\end{equation}

\subsubsection{BiLSTM Network}
LSTM network\cite{hochreiter1997long} encodes the information in one direction $\overrightarrow{h_t}$. However, the BiLSTM network \cite{schuster1997bidirectional} has the ability to encode the sequences from both right $\overrightarrow{h_t}$ as well as left $\overleftarrow{h_{t}}$ contexts. The BiLSTM network computes the forward  $\overrightarrow{h}$ and then backward $\overleftarrow{h}$ hidden states. Finally, the context vector is created by concatenating both hidden states, $\overrightarrow{h_t}$ and $\overleftarrow{h_{t}}$, for the final output $h_{t}=	[ \, \overrightarrow{h_{t}};\overleftarrow{h_{t}} ]\,$ of the network. The processing flow in the BiLSTM network can be expressed as follows: 
\begin{equation}
    \left[\begin{array}{c}
\tilde{{c}}_{t} \\
{f}_{t} \\
{o}_{t} \\
{i}_{t}
\end{array}\right]=\left[\begin{array}{c}
\sigma \\
\sigma \\
\sigma \\
\tanh
\end{array}\right]\left({W}^{\top}\left[\begin{array}{c}
\mathrm{x}_{t} \\
{h}_{t-1}
\end{array}\right]+{b}\right)
\end{equation}
\begin{equation}
    {c}_{t}={i}_{t} \odot \widetilde{c}_{t}+{f}_{t} \odot {c}_{t-1}
\end{equation}
\begin{equation}
    {h}_{t}={o}_{t} \odot \tanh \left({c}_{t}\right)
\end{equation}
where $i_{t}, f_{t},$ and $o_{t}$ indicate the input, forget, and output gates; $\sigma()$ represents the sigmoid function;  ${W}^{\top}$ and $b$ are the trainable parameters; $x_{t}$ is an input vector of the current time step; $\odot$ represents the dot product function.

\subsection{Self-attention}
The limitation of an encoder-decoder model is their fixed-length connection~\cite{liao2019combined} of the intermediate semantic vector. Hence, the encoder compresses the whole sequence of information into a fixed-length vector. Recently, self-attention~\cite{vaswani2017attention} has been widely used in word segmentation~\cite{duan2020attention,cao2018adversarial} to tackle the limitations of encoder-decoder structure to extract as much feature information as possible from the input sequences.  We add a token-level multi-head attention to the output $H=[h_{1}, h_{2}, h_{3} \dots h_{n}]$ of concatenated BiLSTM states to learn the character-level structure of words and capture token-level dependency. The self-attention layer~\cite{vaswani2017attention} before the CRF layer~\cite{jia2019attention} automatically focuses on the specific Sindhi tokens that play a decisive role in word segmentation. We use the scaled dot product to calculate the similarity between query ${Q}$ and key ${K}$ matrices to obtain weights of each word. Then, normalize the obtained score and calculate the weights using softmax. Afterwards, obtain the final attention by a weighted sum of value matrix ${V}$ and the weights. The intuition of an attention mechanism is described as follows:
  \begin{equation}
      \text { attn }({Q}, {K}, {V})=\operatorname{softmax}\left(\frac{{QK}^{T}}{\sqrt{{d}}}\right) {V}
   \end{equation}
where ${Q} \in \mathbb{R}^{n \times 2 d_{h}}, {K} \in \mathbb{R}^{n \times 2 d_{h}}$,  ${V} \in \mathbb{R}^{n \times 2 d_{h}}$, are the query, key and value matrices. $d_{h}$ is the dimension of both $\overrightarrow{h}$ and $\overleftarrow{h}$ hidden units of BiLSTM, which equals to $2{d}_{h}$. Firstly, the attention mechanism linearly maps $\mathbf{Q}$, $\mathbf{K}$ and $\mathbf{V}$ in $h$ times with different weights matrices. Then $h$ projections parallelly  perform scaled dot-product attention. Finally, the results of an attention layer are concatenated and once again mapped to get the new representations. In this paper, we use self-attention as a set $Q = K = V = H$, aims to capture the long dependencies between any two tokens in the input sentence, where $H$ represents the output of the BiLSTM. Formally, the function can be expressed as under:

\begin{equation}
    \text {h}_{i}=\text { attn }\left({Q} {w\,}_{i}^{Q}, {K} {w\,}_{i}^{K}, {V} {w\,}_{i}^{V}\right)
\end{equation}
\begin{equation}
    \mathbf{H}^{\prime}=\left(\text {h}_{i} \mathbin\Vert \ldots \mathbin\Vert \text {head}_{h}\right) \mathbf{w}_{o}
\end{equation}
where ${w\,}_{i}^{Q} \in \mathbb{R}^{2 d_{h} \times d_{k}}$, $ \quad {w\,}_{i}^{K} \in \mathbb{R}^{2 d_{h} \times d_{k}}$, ${w\,}_{i}^{V} \in \mathbb{R}^{2 d_{h} \times d_{k}}$ are the trainable projection parameters and $d_{k}=2 d_{h} / h . \quad {w\,}_{o} \in \mathbb{R}^{2 d_{h} \times 2 d_{h}}$ are the trainable parameters.

% We add a self-attention layer before the CRF classifier, which has the ability to decide how much information to use from token-level components dynamically.  
 %Thus, to enhance the ability of the proposed model, we incorporate self-attention into the BiLSTM network to capture lexical features and semantic information deeply.  It captures the essential semantic information in the input sequences while reducing the focus on less important information.
% \subsection{Word dictionary}
% Character-level representations~\cite{labeau2017character}, and their n-gram based extensions~\cite{shao2017character}, subword-based representations~\cite{bojanowski2017enriching} perform well with respect to learning character-level features from the labeled and unlabelled corpus. However, character-level models lack useful word-level information to determine the character sequences constituting a word. Thus, it is essential for SWS using a word lattice during decoding to use
% word-level information such as a unigram and a bigram. However, this is not necessary for character-based SWS approaches.

\subsection{Output Layer} 
The decoder finds a next-step action conditioned~\cite{zhang2018simple} on historical actions. The overall decoding framework of our SGNWS model is shown by the upper part of Figure \ref{fig:proposed_model}. Following sequence tagging models such as~\cite{huang2015bidirectional,zhang2018simple}. The generated output from the self-attention layer fed to CRF for prediction. Once all $a_{i}$ are generated from the network for
each $x_{i}$, a decoder takes them to predict a sequence
of segmentation labels $\hat{y}=\widehat{y}_{1} \hat{y}_{2} \cdots \hat{y}_{l}$ for $\mathcal{X}$ as, 
\begin{equation}
    \widehat{\mathcal{Y}}=\operatorname{Decoder}(\mathcal{A})
\end{equation}
here $\mathcal{A}={a}_{1} {a}_{2} \cdots {a}_{i} \cdots {a}_{l}$ is output sequence. The \textit{decoder} can be implemented by using softmax or CRF. The softmax can be illustrated as:
\begin{equation}
    \widehat{y}_{i}=\arg \max \frac{\exp \left(a_{i}^{t}\right)}{\sum_{t=1}^{|\mathcal{T}|} \exp \left(a_{i}^{t}\right)}
\end{equation}
where $a_{i}^{t}$ is the value at dimension $t$ in $a_{i}$. 

The CRF \textit{decoder} can be illustrated as:
\begin{equation}
    \widehat{y}_{i}=\arg \max _{y_{i} \in \mathcal{T}} \frac{\exp \left({W}_{c} \cdot {a}_{i}+{b}_{c}\right)}{\sum_{y_{i-1} y_{i}} \exp \left({W}_{c} \cdot {a}_{i}\right)+{b}_{c}}
\end{equation}
where ${W}_{c} \in {R}^{|\mathcal{T}| \times|\mathcal{T}|}$ and ${b}_{c} \in {R}^{|\mathcal{T}|}$ are trainable parameters to model the transition for $y_{i-1}$ to $y_{i}$.

\subsection{Transduction}
The non-segmental multiword tokens identified by the main network are transduced into corresponding components in an additional step. We transduce the identified non-segmental multiword tokens in a context-free fashion. For multiword tokens with two or more valid transductions, we only adopt the most frequent one. We use a generalizing approach to process non-segmental multiword tokens. If there are more than 200 unique multiword tokens in the training set, we train a self-attention-based neural encoder-decoder model equipped with shared BiLSTM. Overall, we utilize context to identify non-segmental multiword tokens and then apply a combination of dictionary and sequence-to-sequence encoder-decoder to transduce them.

\section{Experimental Setup}
\label{sec:experitmental_setup}
This section provides details about the experimental setting of baselines as well as proposed SGNWS models. We use TensorFlow~\cite{abadi2016tensorflow} deep learning framework for the implementation of all neural models on GTX 1080-TITAN GPU to run all the experiments.
\subsection{Datasets}
\label{subsec:data}
We utilize the recently proposed unlabeled  Sindhi text corpus~\cite{ali2019new} in the experimental setting. We convert segmentation into a sequence tagging problem using B, I, E, S, X tagging scheme and split the dataset into $80\%$ for training and $20\%$ for development and test sets. Table \ref{tab:corpus_for_segmentation} shows the complete statistics of the datasets.  

\begin{table}
\centering
\caption{Statistics of the proposed training datasets used in the experiments are presented in the table below. Sent., Tok., UW, and AWL denote the number of sentences, tokens, unique words, and average word length in each dataset, respectively. We concatenate all the domains and represent the combined dataset as SDSEG.}
  \label{tab:corpus_for_segmentation}
  \begin{tabular}{l|c|c|c|c}
    \hline
   Datasets & Sent & Tok & UK  & AWL \\
    \hline
   KN &  24,212 & 601,910 & 10,721 & 3.687\\ 
    \hline
   AA & 19,736 & 521,257 & 14,690 & 3.660\\
    \hline
    WK  & 14,557 & 669,623 & 11,820 & 3.738\\
    \hline
    Tw & 10,752 & 159,130 & 17,379 & 3.820\\
    \hline
     Bk  & 22,496 & 430,923 & 16,127 & 3.684\\
    \hline
    SDSEG & 91,753 & 2,382,843 & 70,737 & 3.717\\
    \hline
\end{tabular} 
\end{table}

\subsection{Baseline Models}
\label{subsubsec:baseline_models}
We conduct several baseline experiments by training the LSTM, BiLSTM using softmax and CRF decoders. We train character-level~\cite{shao2017character} subword representations in baseline experiments to analyze and compare the performance of proposed model. The brief description of each baseline model is described as follows:
 \begin{enumerate}
    \item \textbf{BiLSTM}: Our first baseline utilizes the BiLSTM encoder, which is leveraged with character-level representations, bigrams, and SRL. We employ a softmax classifier in the last layer of the network for decoding tag sequences. The BiLSTM benefits from encoding both forward and backward sequences, efficiently capturing word information at the morphemic level.
    \item \textbf{BiLSTM-CRF}: The second baseline model is built upon a BiLSTM-CRF network with a hyperparameter setting similar to that of the BiLSTM encoder. We utilize character-level representations, bigrams, and SRL. CRF inference is employed in the last layer of the network for decoding purposes.
\end{enumerate}
\subsection{Evaluation metrics}
\label{subsec:evaluation_metric}
We report word boundary precision, recall, and F1-scores ~\cite{shao2017cross}. Precision measures the proportion of correctly predicted word boundaries among all predicted boundaries. Recall measures the proportion of correctly predicted boundaries compared to the actual number of boundaries in the text. F1-score, the harmonic mean of precision and recall, provides a balanced view of these two metrics.

\begin{equation}\label{Eq.precision}
    {\text{Precision}}=\frac{\text{\#(correctly\_predicted\_ tags)}}{\text {\#(predicted tags)}}  \times 100 \%\\
    \end{equation}
    \begin{equation}\label{Eq.recall}
    {\text{Recall}}=\frac{\text{\#(correctly\_predicted\_tags)}}{\text {\#(true\_tags)}}  \times 100 \%\\
    \end{equation}
    \begin{equation}\label{Eq.fscore}
    F1\text{-score}=\frac{2 \times {\text{Precision}} \times {\text{Recall}}}{{\text{Precision}}+{\text{Recall}}} \times 100 \%\\
\end{equation}
\subsection{Parameter setting and training}
\label{subsec:model_training}
The training procedure is to regulate all parameters of the network from training data. We train the baselines and proposed model using the log-likelihood function. The log-likelihood has already been optimized to give strong performances in our baseline experiments compared to global learning~\cite{andor2016globally} to maximize F1-score. We split the dataset into training, development, and test sets. Variational dropout~\cite{gal2016theoretically} of $0.25\%$ is employed to recurrent units. The softmax is used for label classification in baseline models, CRF is added in the last layer of the BiLSTM-CRF and SGNWS models. The Gradient normalization is used to improve the performance~\cite{pascanu2013difficulty}, which re-scales the gradient when the norm goes over a threshold. We use BiLSTM state size of 200 layers, and character embedding size of $64$. We train the models for 40 epochs, and use the Nadam optimizer~\cite{KingmaB14} with initial learning rate of $0.01$ We use beam decoding with beam size of $5$. For each configuration, we trained and evaluated different settings of a manually tuned hyperparameters grid, varying initial learning rate, and input and recurrent dropout rates. All the hyper-parameters for baseline models and SGNWS are kept similar for performance difference and fair comparison. Our SGNWS is implemented by grouping sentences of similar lengths into the same bucket and padding them to the same length. We construct sub-computational graphs for each bucket so that sentences of different lengths are processed more efficiently~\cite{shao2017cross}.

\begin{table}[h]
\centering
\caption{Optimal hyper-parameters for SRL, baselines, and proposed SGNWS model.}
   \label{tab:optimized_hyperparameters}
    \begin{tabular}{l|l|c}
        \hline
      & Hyperparameter & Range\\ 
        \hline
        \multirow{6}{*}{\rotatebox[origin=c]{90}{SRL \quad}}  
    & ${E^{c}}$ representation dimension  & 64\\
     & $E^{b}$ representation dimension  &  64\\
     & $E^{t}$ representation dimension & 64\\
     & ${E^{cbt}}$ representation dimension & 64\\ 
     & Window size & 5\\
     & Subword representation dimension & 100\\
      &  Epochs & 100\\
        \hline
        \multirow{6}{*}{\rotatebox[origin=c]{90}{Neural models \quad}}  
    &  Optimizer & Nadam\\
     & Initial learning rate &  0.01\\
     & Gradient normalization & 5.0\\
     & BiLSTM state size & 200\\
     & Dropout & 0.25\\
     & Decay rate & 0.05\\
     & Batch size & 32\\
     & Epochs & 40\\
      \hline
  \end{tabular}
\end{table}

\section{Results and analysis}
In this section, we firstly report the results of baselines with different configurations on SGSEG benchmark dataset. Then, we describe the validity of the proposed on different datasets as well as SDSEG on the same configurations.  

\subsection{Baseline results}
We investigate the capability of baseline neural models to handle words of different lengths. The table presents the results, demonstrating that the best performances are achieved when the n-gram length is three under all settings. We utilize the BiLSTM encoder with softmax and CRF decoder. The empirical results on the SDSEG dataset are showcased, reporting overall precisions, recalls, and F1-scores. Several observations can be drawn from the results. Table \ref{tab:comparision_of_rnn_variants} illustrates the performance comparison of all models on the SDSEG dataset. Firstly, the LSTM achieves a stable baseline F-Score of 95.29\% on the development and 94.32\% on the test set. The BiLSTM model outperforms both baselines in both the development and test sets due to its bidirectional learning capabilities. However, the BiLSTM-CRF outperforms both BiLSTM baselines, indicating the dominance of CRF over the softmax classifier. Moreover, adding character-level features to the BiLSTM-CRF model surpasses all three baselines. However, BiLSTM-CRF with bigram and trigram-based subword representations yield results close to those of the BiLSTM-CRF+Char model, highlighting the superiority of the character-level approach and the performance gain it provides. We have observed the effects of employing various subword representations.

\begin{table}
\centering
\caption{Neural baselines for Sindhi word segmentation on the SDSEG test set.}
\label{tab:comparision_of_rnn_variants}
   \begin{tabular}{|l|c|c|c|c|}
    \hline
    Model &   P\% & R\% & F1\% \\
    \hline
    BiLSTM-Char &  96.52 & 94.28 & 95.87\\
    \hline
    BiLSTM-Char+bigram &  96.42 & 94.28 & 95.87\\
    \hline
    BiLSTM-SRL & 96.81 & 96.74 & 96.53 \\
    \hline
    BiLSTM-CRF-Char &  96.11 & 95.87 & 96.28 \\
    \hline
    BiLSTM-CRF-Char+bigram &  96.82 & \textbf{97.26} & 96.78 \\
    \hline
    BiLSTM-CRF-SRL  & \textbf{97.17} & \textbf{97.23} & \textbf{97.39}\\
    \hline
\end{tabular}
\end{table}
\begin{figure*}[!h]
\centering
	\includegraphics[width=0.76\textwidth]{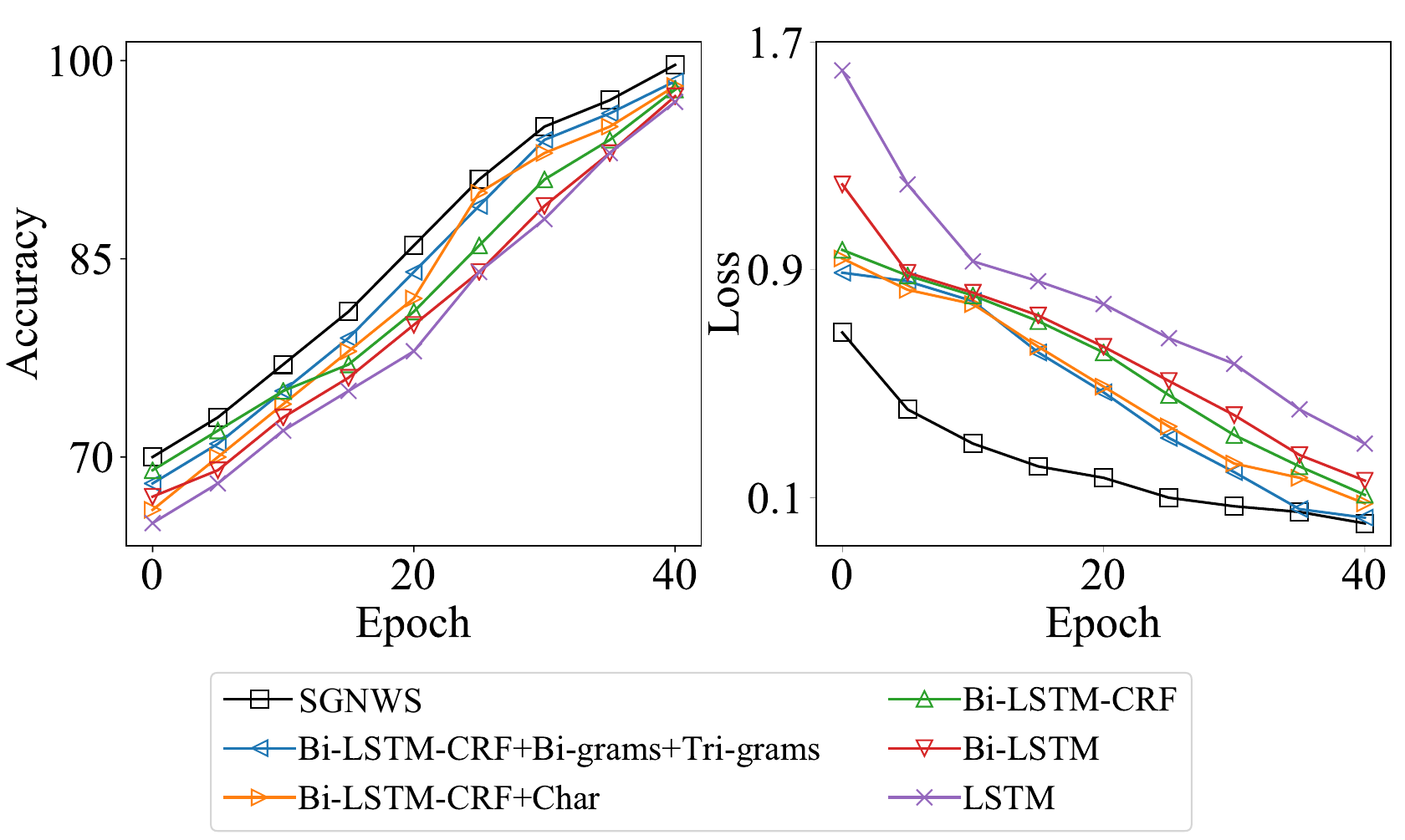}
 	\caption{The training accuracy and loss of baseline and proposed SGNWS models on the SDSEG dataset.}
     \label{fig:training_accurayc_loss}
 \end{figure*}
\subsection{Final results}
The proposed SGNWS model has yielded superior results compared to the baselines, as shown in Table \ref{tab:comparision_of_rnn_variants} for test data. According to the results, SRL proves to be beneficial for the word segmentation task in Sindhi language. The proposed SGNWS model outperforms all the baselines on the SDSEG dataset as well as on five different datasets: Kawish newspaper (KW), Awami-Awaz newspaper (AA), Wikipedia dumps (WK), Twitter (Tw), and books (Bk). Moreover, Figure \ref{fig:training_accurayc_loss} illustrates the training accuracy and loss of the SGNWS model, alongside other models, on the SDSEG dataset. Our proposed SGNWS model surpasses the baselines with a high F1-score of $98.51\%$ on the test set using the SDSEG dataset. This observation indicates that SRL is beneficial in capturing more word boundaries and semantic information for the word segmentation of Sindhi text. The SGNWS model achieves state-of-the-art performance on KW and SDSEG datasets.

% \begin{figure}
% \centering
%     \includegraphics[width=0.48\textwidth]{all_dataset_fscore}
%     \caption{The performance of proposed SGNWS model on the various datasets. The F-Score is reported on the test set of multiple datasets.}
%     \label{fig:performance_comparision_bilstm_crf_subword_on_all_dataset}
% \end{figure}

\begin{table}[h!]
\centering
\caption{Evaluation of the SGNWS model on various datasets. Bold font denote the best results.}
  \label{tab:sgns_all}
  \begin{tabular}{|l|c|c|c|}
    \hline
   Domain  & P\% & R\% & F1\%\\
    \hline
    Kawish News (KN) & \textbf{98.56} & \textbf{98.34}  & \textbf{98.12}\\ 
    \hline
    Awami Awaz (AA) & \textbf{98.17} & 97.67 & 97.69\\
    \hline
    Wiki-dumps (WK) & 97.25 & 96.84 & 96.37\\
    \hline
    Twitter (Tw) & 97.41 & 97.29 & 97.24\\
    \hline
    Books (BK) & 97.57 & \textbf{98.28} & \textbf{97.83}\\
    \hline
    SDSEG & \textbf{98.48} & \textbf{98.64} & \textbf{98.51}\\
    \hline
\end{tabular} 
\end{table}

Table \ref{tab:sgns_all} presents the final results on the test datasets of Kawish, Awami-Awaz, Wikipedia dumps, Twitter, Books, and SDSEG. We present the performances of the SGNWS neural model by utilizing subword representation learning and self-attention. According to the results, it is evident that the model achieved the best performance on the SGSEG dataset, followed by the Kawish and Books datasets. We compare the SGNWS model with the discrete baseline system under a purely supervised setting. The BiLSTM-CRF-CRL model achieves similar performances on the SDSEG dataset compared to the SGNWS model.

\section{Conclusion}
\label{sec:conclusion}
Word segmentation is an essential and non-trivial task in the Sindhi language. While white spaces between words serve as indicators for predicting word boundaries, the presence of space omission and insertion introduces ambiguity into the segmentation process. In this paper, we introduced the SGNWS model, specifically designed to address the challenges associated with Sindhi Word Segmentation (SWS). Our proposed model possesses the capability to automatically learn and extract subword features with minimal language-specific settings. The SGNWS model achieves superior performance in SWS due to its high efficiency and robustness in sequential modeling tasks, enabling it to effectively capture word information at the morphemic level for predicting word boundaries. Empirical results across various unlabeled datasets demonstrate the effectiveness of SGNWS, showcasing state-of-the-art performance across all datasets. Our proposed model serves as an effective and elegant neural solution for SWS, with potential applicability to other sequence tagging problems.

\bibliographystyle{unsrtnat}
\bibliography{references}  %%% Uncomment this line and comment out the ``thebibliography'' section below to use the external .bib file (using bibtex) .





% \end{thebibliography}

\end{document}